\documentclass{article} 
\usepackage{collas2024_conference,times}
\usepackage{easyReview}

\usepackage{amsmath,amsfonts,bm}

\def\eqref#1{equation~\ref{#1}}

\def\1{\bm{1}}

\def\vv{{\bm{v}}}

\DeclareMathAlphabet{\mathsfit}{\encodingdefault}{\sfdefault}{m}{sl}
\SetMathAlphabet{\mathsfit}{bold}{\encodingdefault}{\sfdefault}{bx}{n}

\usepackage{hyperref}
\usepackage{todonotes}
\hypersetup{
    colorlinks=true,
    linkcolor=red,
    filecolor=magenta,
    urlcolor=blue,
    citecolor=purple,
    pdftitle={Overleaf Example},
    pdfpagemode=FullScreen,
    }
\usepackage{amsmath,amssymb,amsfonts}
\usepackage{algorithm}
\usepackage{algorithmic}
\usepackage{float}
\usepackage{multirow}
\usepackage[f]{esvect}
\usepackage{adjustbox}
\usepackage{subcaption} 
\usepackage{wrapfig}
\usepackage{graphicx} 
\providecommand{\keywords}[1]{\textbf{Keywords:} #1}

\title{Learning to learn without forgetting using attention}

\author{Anna Vettoruzzo  \\
Center for Applied Intelligent Systems Research (CAISR) \\
Halmstad University \\
Sweden \\
\texttt{anna.vettoruzzo@hh.se} \\
\And 
Joaquin Vanschoren  \\
Automated Machine Learning Group \hspace{6em}  \\
Eindhoven University of Technology \\
Netherlands \\
\texttt{j.vanschoren@tue.nl} \\
\And 
 Mohamed-Rafik Bouguelia  \\
Center for Applied Intelligent Systems Research (CAISR) \\
Halmstad University \\
Sweden \\
\texttt{mohamed-rafik.bouguelia@hh.se} \\
\And 
Thorsteinn Rögnvaldsson  \\
Center for Applied Intelligent Systems Research (CAISR) \\
Halmstad University \\
Sweden \\
\texttt{thorsteinn.rognvaldsson@hh.se} \\
}

\collasfinalcopy 

\begin{document}

\maketitle

\begin{abstract}
Continual learning (CL) refers to the ability to continually learn over time by accommodating new knowledge while retaining previously learned experience. While this concept is inherent in human learning, current machine learning methods are highly prone to overwrite previously learned patterns and thus forget past experience. Instead, model parameters should be updated selectively and carefully, avoiding unnecessary forgetting while optimally leveraging previously learned patterns to accelerate future learning. Since hand-crafting effective update mechanisms is difficult, we propose meta-learning a transformer-based optimizer to enhance CL. This meta-learned optimizer uses attention to learn the complex relationships between model parameters across a stream of tasks, and is designed to generate effective weight updates for the current task while preventing catastrophic forgetting on previously encountered tasks. Evaluations on benchmark datasets like SplitMNIST, RotatedMNIST, and SplitCIFAR-100 affirm the efficacy of the proposed approach in terms of both forward and backward transfer, even on small sets of labeled data, highlighting the advantages of integrating a meta-learned optimizer within the continual learning framework.
\end{abstract}

\section{Introduction}
Continual learning (CL), or lifelong learning, refers to the ability of a machine learning system to continuously acquire knowledge from a stream of learning tasks while retaining previously learned experience \citep{cossu2021continual}. 
CL involves a non-stationary stream of tasks where the data distribution evolves over time \citep{son2023meta}.
For instance, consider an autonomous vehicle driving across different countries under varying weather conditions and road surfaces. The vehicle must constantly adapt to changes in the environment by leveraging past knowledge 
to accelerate the learning process without forgetting how to handle past environments.
Conventional neural networks will update all model parameters (weights) when adapting to new tasks, overwriting previously stored information, and therefore tend to gradually forget previously learned knowledge, a phenomenon known as catastrophic forgetting. While one could store samples from all previously encountered tasks and use them to repeatedly retrain the model, this is very inefficient. Instead, model parameters should be updated selectively and carefully, protecting certain parameters to avoid forgetting while allowing others to adapt quickly to accelerate future learning.

Current CL techniques do this via regularization based on hand-crafted heuristics, e.g., by protecting weights that had a significant impact in previous tasks \citep{kirkpatrick2017overcoming}, by keeping an episodic memory with examples from previous tasks (which has scalability and privacy issues), or by gradually extending the architecture with new parameters (which limits scalability).
Meta-learning methods \citep{schmidhuber1987evolutionary, santoro2016meta, finn2017model, vettoruzzo2023advances} can be used to adapt to new tasks very efficiently, e.g., by meta-learning a gradient descent optimizer, but prior works typically ignores the sequentiality and non-stationarity of CL problems. Finally, sparsification techniques allow learning a subnetwork for every single task, thereby reducing interference across diverse tasks while promoting information sharing among related tasks \citep{kang22b, sokar2021spacenet}. However, determining the optimal sparsity level can be challenging and the model's capacity may quickly be surpassed on longer streams of tasks.

This paper presents a novel approach, integrating meta-learning into the CL framework. It introduces a transformer-based meta-optimizer network that learns how to adapt each parameter of a separate classifier network using attention to learn the complex relationships between the parameters across a lifelong stream of tasks. The meta-optimizer is trained to predict task-specific weight updates for fast adaptation to new tasks while minimizing interference with previous tasks to avoid forgetting and scales to long streams of tasks. Additionally, the proposed approach ensures robust performance on future tasks while leveraging only a small set of labeled data, thanks to the generalization capability of meta-learning. This appears to be particularly useful in real-world applications where new tasks are regularly encountered, and the model needs to generalize to them. The idea of this paper draws inspiration from the presence of task-specific pathways within the classifier's weights \citep{pathways}. The meta-optimizer learns which sets of weights are important for particular tasks, and avoids overwriting pre-existing knowledge already stored in these weights. Additionally, by identifying overlapping pathways for related tasks, it also has the potential to optimize shared weights and increase performance on previously seen tasks (backward transfer). 

\begin{figure}[tbp]
\centering
\subfloat[][Proposed approach] 
{\includegraphics[width=.46\textwidth]{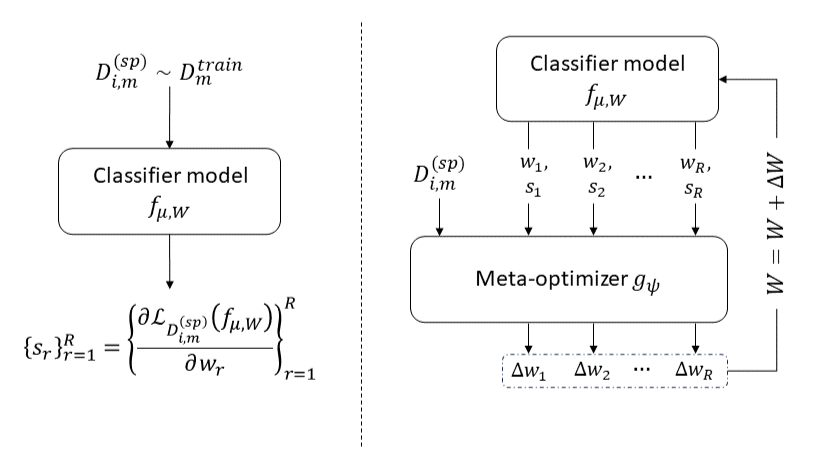}\label{fig:proposed}} \qquad
\subfloat[][Meta-optimizer]
{\includegraphics[width=.34\textwidth]{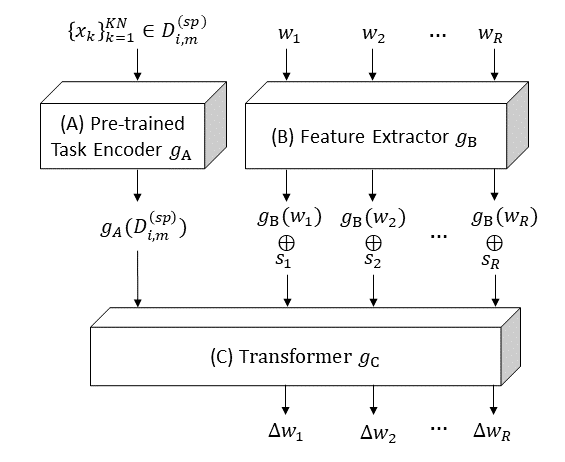}\label{fig:optimizer}}
\caption{(a) Overall framework of the proposed approach. The support set of the current data batch $D_{i,m}^{(sp)}$ is fed into the classifier model $f_{\mu, W}$ to derive importance scores $\{s_r\}_{r=1}^R$. These scores, along with the weights of the classifier model that we want to optimize $W = \{w_r\}_{r=1}^R$, and $D_{i,m}^{(sp)}$, serve as inputs to the meta-optimizer $g_\psi$ for predicting weight updates tailored to task $\mathcal{T}_m$ from which the data are sampled. (b) Architecture of the meta-optimizer. (A) The pre-trained task encoder learns a vector representation that characterizes the current task. (B) The feature extractor maps weight values into the feature space. (C) The transformer encoder predicts the weight updates.}
\label{fig:approach}
\end{figure}

To achieve this, we need to provide information about the current task and weights to the transformer-based optimizer. As illustrated in Figure \ref{fig:approach}, we first compute a task-specific \textit{importance score} for each weight based on the gradient of the loss function when feeding labeled data from the current task to the classifier model before training on that task. By zeroing out the importance scores below a certain threshold, the meta-optimizer is forced to optimize the subset of weights considered important for the current task, thereby preventing catastrophic forgetting. The meta-optimizer takes as input a small set of labeled data from the current task, the current weights of the classifier model, and the importance scores, and outputs the weight updates in sequence, taking into account the specific task and the values of all previous weights. Internally, a pre-trained task embedding is used to allow the transformer to take into account the similarities between the current and previous tasks. 

We demonstrate the ability of our meta-optimizer to predict effective weight updates on various datasets within the CL framework, even with small sets of labeled data. This proves particularly useful when the inference process needs to be performed effectively after observing only a limited amount of labeled data, a common scenario in real-world applications. Moreover, we find that predicting only the weight updates of the last layers of the classifier is sufficient to achieve good results in terms of backward transfer, supporting the idea that the last layers encode more task-specific information compared to earlier layers \citep{ramasesh2020anatomy}. Importantly, this approach also eliminates the need to select optimizer-related hyperparameters for the classifier network, such as learning rate, weight decay, and optimizer type. Instead, the hyperparameters of the meta-optimizer proposed in this paper are selected only once at the beginning of the training, and the model itself automatically learns which optimization strategy is best suited for each task.
This is especially important in the CL setting, where the environment is constantly evolving, and hyperparameters that work well for a given task may not be effective for future ones. 
To our knowledge, this work represents the first approach proposing the use of a transformer model as a meta-optimizer applied in the context of CL.
This establishes a foundation for future research to explore the potential of transformers in optimizing neural networks for CL.

In the remainder of this paper, we'll discuss the relationships with prior work in Sect.~\ref{sec:relatedworks}, and preliminaries in Sect.~\ref{sec:preliminaries}, before detailing our method in Sect.~\ref{sec:approach}. We will present our experiments in Sect.~\ref{sec:experiments} and conclusions in Sect.~\ref{sec:conclusions}.

\section{Related work} \label{sec:relatedworks}
The CL problem has a number of variants, depending on whether the task identity is available at test time, i.e., task-incremental learning (TIL), or whether it has to be inferred, in which case one may have to predict new classes, i.e., class-incremental learning (CIL), or the same classes across domains, i.e., domain-incremental learning (DIL) \citep{van2019three, wang2023comprehensive}.
We can classify CL methods by the strategy used to alleviate catastrophic forgetting. Memory-based methods, such as those proposed in \cite{buzzega2020dark, rebuffi2017icarl, lopez2017gradient, lopez2017gradient2}, maintain a buffer of past data for knowledge rehearsal during inference. Regularization-based methods, as explored in \cite{kirkpatrick2017overcoming, serra2018overcoming, aljundi2018selfless, zenke2017continual}, impose regularization on network parameters with diverse heuristics during training. Meanwhile, dynamic architectural methods, as presented in \cite{sokar2021spacenet, hemati2023partial, vladymyrov2023continual, kang22b, aljundi2017expert, serra2018overcoming, golkar2019continual, rusu2016progressive}, allocate distinct sets of parameters for each task.
Related to the last category, one way to completely avoid interference among diverse tasks is to grow new neural network branches for each new task while freezing the parameters of previous tasks \citep{kang22b} or dedicating a model copy to each task \citep{aljundi2017expert}. However, these methods lead to high memory requirements as the number of tasks increases. Alternative approaches maintain a constant architecture size, allocating only a portion of the network to each task, where neurons are selected based on the concept of network sparsity \citep{sokar2021spacenet}, self-attention based masks \citep{serra2018overcoming} or neuron-specific metrics \citep{golkar2019continual}. A related idea is to apply hypernetworks \citep{hypernetworks} combined with regularization and generative replay techniques to generate task-specific network parameters and mitigate forgetting \citep{von2019continual}. Since generating all weights is computationally expensive for large architectures, partial hypernetworks suggest generating weights only for the final layers of the classifier while keeping the initial layers unchanged \citep{hemati2023partial, vladymyrov2023continual}. 

In meta-learning \citep{vettoruzzo2023advances}, the goal is to train a model on a series of tasks so that a new task can be quickly learned with only a small set of labeled data, for instance by finding an initialization of the model parameters that allows fine-tuning on a small set of labeled data from that task \citep{finn2017model}, or by meta-learning the optimization rule that is at the basis of the neural network training \citep{chen2022learning}. Instead of manually choosing an optimization strategy (such as \cite{tseng1998incremental, riedmiller1993direct, duchi2011adaptive, kingma2014adam}), \cite{andrychowicz2016learning, ravi2016optimization, chen2017learning} propose an LSTM-based meta-learner that is directly trained to optimize a classifier model.
Similarly, \cite{wichrowska2017learned, metz2022velo, irie2023automating} and \cite{jain2024mnemosyne} explore more intricate model architectures to achieve the same objective through advanced and complex structures. In particular, \cite{jain2024mnemosyne} introduces a novel class of learnable optimizers based on spatio-temporal attention transformers, aiming to train large and intricate architectures like transformers and match state-of-the-art hand-designed optimizers. Differently from \cite{jain2024mnemosyne}, the primary focus of this paper is to introduce a method that learns to selectively update the parameters of a classifier network for each task in a sequential training stream, adjusting the optimization strategy accordingly while retaining the previous knowledge stored in the classifier's weights.
Indeed, none of the previous approaches work well in this CL scenario, as they are primarily designed for cases where tasks are presented simultaneously. This results in a tendency to gradually forget the previous knowledge when new tasks are added to the stream and old tasks cannot be revisited. To address this challenge, La-MAML \citep{gupta2020look} proposes a solution that mitigates catastrophic forgetting by incorporating a replay buffer and optimizing a meta-objective. MER \citep{riemer2018learning}, inspired by GEM \citep{lopez2017gradient}, integrates an experience replay module to maximize the transfer of information and minimize interference between tasks. In contrast, OML \citep{javed2019meta} meta-learns an optimal representation offline and subsequently freezes it for continual learning on new tasks. Although these methods demonstrate notable performance on CL datasets, they require a large memory to store previous examples in the replay buffer, and they cannot generalize to training streams where the data distribution evolves significantly. Furthermore, they are not designed to deal with tasks where only a small set of labeled data is available at test time. 

Our contribution involves designing an approach that replaces hand-crafted heuristics with meta-learning to adapt effectively and efficiently (in terms of labeled data) to streaming tasks without forgetting past knowledge. The proposed method directly optimizes the weights of a neural network classifier, even on small data samples, prevents interference with existing knowledge, doesn't require a memory buffer, and scales to long streams of tasks. It ensures a balance between the ability to preserve past knowledge and the ability to learn new tasks - an aspect commonly referred to as the stability-plasticity trade-off \citep{grossberg2012studies}.

\section{Preliminaries and notation} \label{sec:preliminaries}

\subsection{Meta-learning} \label{subsec:meta-learning}
Meta-learning has recently emerged as a powerful paradigm to enable a learner to effectively learn unseen tasks with only a few samples by leveraging prior knowledge learned from related tasks \citep{vettoruzzo2023advances}. In the meta-training phase a set of training tasks $\{\mathcal{T}_i\}_{i=1}^{T}$ is sampled from a task distribution $p(\mathcal{T})$. Each task corresponds to data generating distributions $\mathcal{T}_i \triangleq \{ p_i(x), p_i(y \vert x) \}$, and the data sampled from each task is divided into a \emph{support set}, $D_i^{(sp)}$, containing $K$ training examples per class, and a \emph{query set}, $D_i^{(qr)}$. During the meta-test phase, a completely new task $\mathcal{T}_{new}$ with few labeled data $D_{new}^{(sp)} \triangleq \{x_k, y_k\}_{k=1}^K$ is observed and the model is fine-tuned on $D_{new}^{(sp)}$ to achieve good generalization performance on new unlabeled test examples from $\mathcal{T}_{new}$. Optimization-based methods typically frame meta-learning as a bi-level optimization problem. At the inner lever, a neural network $f_\theta$ produces task-specific parameters $\theta'_i$ using $D_i^{(sp)}$, while at the outer level, the initial set of meta-parameters $\theta$ is updated by optimizing the performance of $f_{\theta'_i}$ on the query set of the same task.
The result is a model initialization $\theta$ that can effectively be adapted to new tasks using only a few ($K$) training examples and a few gradient updates.

\subsection{Continual learning} \label{subsec:cl}
Continual learning (CL) is characterized by learning from evolving data distributions. In practice, training samples from different distributions arrive sequentially, requiring a CL model to learn new tasks incrementally. Moreover, the model must retain the previous knowledge and perform well on the test set of all tasks seen so far, with limited or no access to samples from previous tasks. Therefore, in the context of CL, data is presented as a non-stationary \emph{stream} of tasks, defined as $\vv{\mathcal{T}} = \{\mathcal{T}_{m}\}_{m=1}^M$. At each stage $m$, the entire dataset of task $\mathcal{T}_{m}$ is available for updating the model. The data sampled from each task is split into a training set $D_m^{train}$ and a test set $D_m^{test}$. A label $t_m$ identifies the current task and may or may not be available, depending on the categorization presented in Section \ref{sec:relatedworks}. By combining meta-learning with continual learning (a scenario defined as meta-continual learning (MCL) in \cite{son2023meta}), different batches of data can be sampled from $D_m^{train}$ during training. Each batch is then divided into a limited support set $D_{i,m}^{(sp)}$, used to adapt the model to the current task, and a query set $D_{i,m}^{(qr)}$, to evaluate the performance of the adapted model, as described in Section \ref{subsec:meta-learning}. As the distribution evolves over time, different tasks are characterized by different data distributions, i.e., $p_m(x, y) \neq p_{m+1}(x, y)$, which implies that either the input distribution, the output distribution, or both may differ between different experiences. 
At inference time, when any task $\mathcal{T}_{n}$ (which can potentially be one of the earliest or the future ones) is sampled from the stream, only a small set of labeled data sampled from $D_n^{test}$ is available. The model can adapt to it and then make accurate predictions on new unlabeled test examples in $D_n^{test}$.
Consequently, the main challenge in CL is to mitigate catastrophic forgetting on previously encountered tasks while continually acquiring new knowledge from new tasks \citep{son2023meta}.

\section{Proposed approach} \label{sec:approach}
In this section, we provide a detailed description of the proposed approach. The main objective is to meta-learn an optimizer that dynamically updates the weights of a classifier network, as illustrated in Figure \ref{fig:proposed}. 
The goal is to customize the optimization process for each task by learning to optimize a specific set of weights, thereby minimizing interference with previously acquired knowledge while promoting knowledge transfer across related tasks. Additionally, we enhance the generalization capability of our proposed approach by leveraging meta-learning algorithms to enable efficient and effective adaptation to each task without necessitating an identifier for the current task (defined as $t_m$ in Section \ref{subsec:cl}) and without the requirement of a replay buffer to store examples from previous tasks.
In summary, our adjoint network leverages task-specific pathways within the classifier's weights, which allows for a more nuanced and adaptable approach to weight updates. This ensures that new knowledge can be incorporated without indiscriminately overwriting existing knowledge, directly addressing the challenge of catastrophic forgetting. Thanks to meta-learning, the model also guarantees to quickly adapt to new tasks by transferring relevant knowledge to it.

As depicted in Figure \ref{fig:proposed}, the meta-optimizer $g_\psi$, embodied as a transformer model, aims to predict weight updates for the classifier $f_{\mu, W}$, which is parameterized by two sets of parameters. Here, $\mu$ represents the meta-learned parameters that aim to enhance generalizability across different tasks, while $W=\{\textnormal{w}_r\}_{r=1}^R$ consists of all weights optimized using the output of the meta-optimizer. For the sake of simplicity, we will refer to $\theta$ as the set of parameters that are meta-learned, comprising both $\psi$ and $\mu,$ and to $W$ as the classifier weights that are directly optimized by the meta-optimizer instead of being \emph{trained} with a fixed optimizer.
For a given task $\mathcal{T}_m$, $\theta$ is adapted to $\theta'$ based on batches of data from that task so that the produced $W$ achieves good predictions for the task at hand. Therefore, we can formulate this as an optimization problem where the objective is as follows:
\begin{equation}
    \min_{\theta} \sum_{D_{i,m}} \mathcal{L}(\{\mu \cup \underbrace{g_\psi(D_{i,m}^{(sp)}}_{W}\}), D_{i,m}^{(qr)})
\end{equation}
where $D_{i,m}, i=1,...,\mathit{I}$ are the data batches sampled from the dataset $D_m$ associated with task $\mathcal{T}_m$.
Ideally, all weights of the classifier model can be part of $W$ except for those in the last layer (i.e., classifier head), which are always meta-learned to ensure adaptation to the $N$ classes of the current task. Indeed, the classifier head in the proposed approach maintains a consistent structure (i.e., $N$ output units) throughout all the tasks, neither changing (as in TIL) nor expanding to accommodate additional classes with each new task (as in CIL). Therefore, the classifier head is similar to the one used in DIL approaches, with the main difference being that it can be adapted to classify different classes thanks to meta-learning. Indeed, the primary goal of the classifier
model is to distill informative features that characterize the current task in the deeper layers and enable accurate classification within the $N$ classes after the adaptation of the classifier head to the current task. 
For this reason, our approach stands between TIL and CIL methods. Unlike TIL, it does not require a label identifier for the current task, and contrary to CIL, it does not aim to expand the classifier to include all classes encountered so far. Rather, our approach aligns more closely with the meta-learning framework detailed in Section \ref{subsec:meta-learning}. Each task in the training stream consists of data batches, each featuring a small support set and a query set. The support set is used to adapt the model to the specific task, enabling classifications within the $N$ classes present in the current task, and the query set is used to validate the adapted model and update the meta-learned parameters.
We present the architecture of the meta-optimizer in Section \ref{subsec:metaoptim}, and we provide the complete algorithm in Section \ref{subsec:algo}.

\subsection{Meta-optimizer} \label{subsec:metaoptim}
As illustrated in Figure \ref{fig:optimizer}, the meta-optimizer comprises three key components: (A) a pre-trained task encoder $g_A$, (B) a feature extractor $g_B$, and (C) a transformer encoder model $g_C$. For each data batch $D_{i,m}$ sampled from $\mathcal{T}_m$, the pre-trained task encoder $g_A$ takes as input the data from the support set of the current batch $D_{i,m}^{(sp)}$ without the corresponding labels. This data is denoted as $\{x_{k}\}_{k=1}^{KN} \in D_{i,m}^{(sp)}$, where $K$ examples are available for each of the $N$ classes. After extracting meaningful features from the data, an averaging layer computes the mean of the transformed data along the $KN$ examples, producing a vector representation $g_A(D_{i,m}^{(sp)})$ that characterizes the current task. This representation is invariant to permutations of the examples and is expected to capture task-specific information. 
The feature extractor $g_B$ is primarily responsible for mapping weight values into a size suitable for the transformer model. 
For convolutional weights $\{\textnormal{w}_r\}_{r=1}^R$, with dimension $n_{input} \times n_{output} \times k \times k$, two encoding approaches are employed: (a) \emph{sequential allocation}, where weights are flattened into a one-dimensional vector of size $(n_{input} \cdot n_{output} \cdot k \cdot k$, 1), and (b) \emph{spatial allocation} which generates $n_{input} \cdot n_{output}$ vectors with size $k^2$. The importance scores are modified in a similar manner: for \emph{sequential allocation}, they are flattened into a vector of size $n_{input} \cdot n_{output} \cdot k \cdot k$, while for \emph{spatial allocation}, an average value for each kernel is computed by averaging the $k^2$ scores and squeezing the vector into a size $(n_{input} \cdot n_{output}, k^2)$.

The output from the feature extractor is concatenated with the importance score vector before being fed into the transformer model. This operation is denoted as $g_B(\textnormal{w}_r) \oplus s_r$ for $r=1, \cdots, R$ in Figure \ref{fig:optimizer}. It is worth noting that both the output of the pre-trained task encoder and the output of the feature extractor, once concatenated with the importance score, have the same dimensionality. Finally, the transformer encoder $g_C$ learns weight updates specific to the current task, i.e., $\{\Delta \textnormal{w}_r\}_{r=1}^R$. The transformer mainly consists of a learnable positional encoding, one or more encoder layers, and a linear layer on top that generates the corresponding update for the input weight vector. 

In essence, the meta-optimizer operates as a cohesive unit, extracting task-specific information, effectively mapping weight values, and generating task-specific weight updates through the transformer encoder. This entire model is meta-trained to enhance its generalization capability across different tasks, ensuring effective learning with only a limited number of examples.

\subsection{Algorithm} \label{subsec:algo}
As highlighted before, $\theta$ refers to the set of parameters that are meta-learned and $W$ to the classifier weights that are directly optimized by the meta-optimizer. To further emphasize this distinction, we use the notation $\theta'$ to indicate the parameters optimized by a fixed optimizer and $\tilde{W}$ to indicate the weights updated by the meta-optimizer.
It is important to notice that $\theta$ comprises both the parameters of the meta-optimizer, denoted as $\psi$, and some parameters of the classifier model, i.e., $\mu$. Indeed, as previously mentioned, the last layer of the classifier model is always meta-learned since it must be adapted to the classes of the current task. The complete algorithm of the proposed approach is presented in Algorithms \ref{algo1} and \ref{algo2}. 

At each stage $m$, a task $\mathcal{T}_m$ is sampled from the training stream $\vv{\mathcal{T}}$ and the whole model is trained to learn such task effectively, without forgetting the previously encountered ones. As mentioned in Section \ref{subsec:cl}, the data from each task is split into $D_m^{train}$ and $D_m^{test}$. In line 6, batches of data $D_{i,m}, i=1, ..., \mathit{I}$ are sampled from $D_m^{train}$ and each batch is split into two disjoint sets, $D_{i,m}^{(sp)}$ and $D_{i,m}^{(qr)}$. The support set is used in line 9 to \emph{Adapt\&Optimize} the classifier model for the current task. This function 
is presented in detail in Algorithm \ref{algo2}. After a first initialization step (lines 1 and 2 in Algortihm \ref{algo2}), $Q$ iterations are performed to adapt the meta-learned parameters $\theta'$ to the current task. For each iteration, an importance score is computed in line 5 of Algorithm \ref{algo2} for each weight $\{\textnormal{w}_r\}_{r=1}^R$ in $W$. 
This calculation is performed by inputing $D_{i,m}^{(sp)}$ in the classifier model $f_{\mu', W}$ and computing the gradient of the loss function $\nabla_W \mathcal{L}_{D_{i,m}^{(sp)}}(f_{\mu', W})$ before the weights $W$ are optimized for the current data batch. These gradients are referred to as importance scores $\{s_r\}_{r=1}^R$. They provide information about the importance of optimizing each weight for making predictions on $D_{i,m}$, and more generally on $\mathcal{T}_{m}$, and encourage the meta-optimizer to learn an update strategy similar to gradient descent. Scores below a certain threshold are zeroed out since they do not serve in solving the task at hand. This guarantees that we don't overwrite knowledge from previous tasks, thus avoiding forgetting.

\begin{minipage}[t]{0.52\textwidth}
\begin{algorithm}[H]
\caption{Meta-training from a stream of tasks}\label{algo1}
\textbf{Require} Stream of tasks $\vv{\mathcal{T}}$, inner steps $Q$, learning rates $\alpha, \beta$
\begin{algorithmic}[1]
\STATE Randomly initialize $\theta = \{\mu, \psi\}$ and $W$
\FOR {$m= 1, \cdots, M$}
    \STATE Sample $\mathcal{T}_m$ from $\vv{\mathcal{T}}$
    \STATE Sample data $D_m^{train}$ from $\mathcal{T}_m$
    \WHILE{not done}
        \STATE Get data batches $D_{i,m}, i=1,...,\mathit{I}$ from $D_m^{train}$
        \FOR {\textbf{all} $D_{i,m}$}
        \STATE Split data $D_{i,m}=\{D_{i,m}^{(sp)}, D_{i,m}^{(qr)}\}$ 
        \STATE $\theta'_i, \tilde{W}_i \gets$ \emph{Adapt\&Optimize}$(Q, \alpha, D_{i,m}^{(sp)}, \theta, W)$
        \STATE $\{\mu'_i, \psi'_i\} \leftarrow \theta'_i$
        \ENDFOR
        \STATE Update $\theta \gets \theta - \beta \nabla_{\theta} $ $\sum_{D_{i,m}}$ $\mathcal{L}_{D_{i,m}^{(qr)}}(f_{\mu'_i, \tilde{W}_i})$ \label{line:update}
        \STATE Set $W \gets \tilde{W}_{\mathit{I}}$
    \ENDWHILE
\ENDFOR
\STATE \textbf{return} $\theta, W$
\end{algorithmic}
\end{algorithm}
\end{minipage}
\hfill
\begin{minipage}[t]{0.45\textwidth}
\begin{algorithm}[H]
\caption{\emph{Adapt\&Optimize}}\label{algo2}
\textbf{Require} Adaptation steps $Q$, learning rate $\alpha$, support set $D_{i,m}^{(sp)}$, meta-learned parameters $\theta=\{\mu, \psi\}$, optimized parameters $\tilde{W}$
\begin{algorithmic}[1]
\STATE Initialize $\mu' \gets \mu, \psi' \gets \psi$ 
\STATE Set $\theta' = \{\mu', \psi'\}$
\FOR {$q= 1, \cdots, Q$}
    \STATE Initialize $W \gets \tilde{W}$
    \STATE Compute importance scores \par
    \hskip\algorithmicindent $\{s_r\}_{r=1}^R \gets \nabla_W \mathcal{L}_{D_{i,m}^{(sp)}} (f_{\mu', W})$
    \STATE $\Delta W \gets g_{\psi'}(D_{i,m}^{(sp)}, W, \{s_r\}_{r=1}^R)$
    \STATE $\tilde{W} \gets W + \Delta W$
    \STATE $\theta' \gets \theta'-\alpha \nabla_{\theta'} \mathcal{L}_{D_{i,m}^{(sp)}}(f_{\mu', \tilde{W}})$ \label{line:adapt}
\ENDFOR
\STATE \textbf{return} $\theta', \tilde{W}$
\end{algorithmic}
\end{algorithm}
\end{minipage}

In line 6 of Algorithm \ref{algo2}, the resulting scores $\{s_r\}_{r=1}^R$, together with $W$ and $D_{i,m}^{(sp)}$ are input in the meta-optimizer $g_{\psi'}$ which predicts the weight updates $\Delta W$, as described in Section \ref{subsec:metaoptim}. The updates are then summed to the current weight values in line 7, resulting in the optimized weights $\tilde{W}$. These weights are then used to adapt $\theta'$ to the specific task. This adaptation is performed using the optimized weights $\tilde{W}$ and the batch of data sampled from the current task, i.e., $D_{i,m}^{(sp)}$.
At the end of this process, the adapted parameters $\theta' = \{\mu', \psi'\}$ and the optimized weights $\tilde{W}$ are returned and used in line 12 of Algorithm \ref{algo1} to update the original $\theta$. To do so, the post-adaption loss is computed with the query set $D_{i,m}^{(qr)}$ and the classifier model $f_{\mu'_i, \tilde{W}_i}$, where $\mu'_i$ are the parameters adapted to the support set of the current batch $D_{i,m}^{(sp)}$ and $\tilde{W}_i$ are optimized with the meta-optimizer. Here, any optimizer of choice, such as Adam, can be used to update the parameters $\theta$.
The result of meta-training (line 16) is a set of parameters $\theta = \{\mu, \psi\}$ that generalize well for all tasks. Meanwhile, the parameters $W$ are optimized in a way that enables them to solve the current task effectively without overwriting the knowledge already encoded from earlier tasks.

During inference, the model is given a small set of labeled data from a task $\mathcal{T}_n \in \vv{\mathcal{T}}$ which may correspond to one of the previous or future (new) tasks. 
More specifically, a small set of labeled data $D_{test,n}^{(sp)}$ is sampled from the test set $D_n^{test}$ of a task $\mathcal{T}_n$. The parameters $\theta$ and $W$ are then adapted and optimized as outlined in Algorithm \ref{algo2}, possibly with a different number of adaptation steps $Q$ than the one used for training. The model, now parameterized by $\theta'_n$ and $\tilde{W}_n$, can be used to make accurate predictions on new unlabeled test examples from $D_n^{test}$. 

\section{Experiments} \label{sec:experiments}

\subsection{Datasets and benchmark methods} \label{sec:ds_and_methods}
The effectiveness of the proposed approach is evaluated across common CL datasets, including SplitMNIST, RotatedMNIST, and SplitCIFAR-100. The selection of these datasets is primarily guided by their extensive usage in CL literature and the intention to assess the effectiveness of the proposed approach in DIL scenarios by presenting results on the RotatedMNIST dataset. SplitMNIST is a variation of the original MNIST dataset, comprising five streaming tasks, each featuring two consecutive MNIST digits. RotatedMNIST is generated by rotating MNIST images by a random degree rotation between 0 and 180, consisting of a total of 10 tasks. This dataset is usually used in domain-incremental learning, where the structure of the problem is always the same (classifying digits from 0 to 9), but the input distribution changes due to domain shifts. Similarly to SplitMNIST, SplitCIFAR-100 is constructed by selecting five classes per task from the CIFAR-100 dataset, and creating a stream with five distinct tasks.

Results are compared with several established CL methods. Given that the proposed approach lies between TIL and CIL, as discussed in Section \ref{sec:approach}, a diverse set of baselines is used for comparison. For TIL, this includes regularization-based strategies like EWC \citep{kirkpatrick2017overcoming} and SI \citep{zenke2017continual}, approaches that integrate meta-learning with CL such as La-MAML \citep{gupta2020look} and Sparse-MAML \citep{von2021learning}, as well as dynamic architectural methods like WSN \citep{kang22b}. Furthermore, two CIL methods, iCaRL \citep{rebuffi2017icarl} and DER++ \citep{buzzega2020dark}, are also included in the comparison.

For all methods, three metrics are used for performance evaluation. The average accuracy (``Avg accuracy") represents the mean classification accuracy across all tasks. Backward transfer (BWT) \citep{lopez2017gradient} measures the ability to retain past knowledge from previous tasks. Negative BWT occurs when learning a task results in decreased performance on some preceding tasks, indicating catastrophic forgetting. Forward transfer (FWT), on the other hand, measures the ability to transfer knowledge to learn a new task that has never been encountered before. 

\subsection{Training setting}
To ensure fair and reliable comparisons, all baselines use similar hyperparameters and model architectures. The default setting is used for the value of the specific parameters for the baseline methods. All models are trained for $10^5$ steps, with a batch size of $32$ and $K=5$, where $K$ denotes the number of training examples per class in the support set of each data batch. 
The importance scores are normalized within the range $[0,1]$ and only the top $60\%$ are retained, while the rest are zeroed out, as discussed in Section \ref{subsec:algo}. Since negligible differences (less than $0.1\%$) were observed when comparing \emph{sequential allocation} with \emph{spatial allocation} of the weights, the latter is chosen for all experiments to more efficiently utilize GPU memory. Two distinct classifier architectures are used for SplitMNIST/RotatedMNIST and SplitCIFAR-100. The first classifier includes a convolutional layer with 32 filters and a classifier head, while the second model comprises three convolutional layers with 100 filters and concludes with a classifier head.
The meta-optimizer, as described in Section \ref{subsec:metaoptim}, consists of three parts. The task encoder (part A) is pre-trained on the SVHN dataset for SplitMNIST and RotatedMNIST, and on ImageNet for SplitCIFAR-100. The feature extractor (part B) comprises one hidden layer with 16 neurons, while the transformer architecture (part C) is a transformer encoder with two layers, GeLU activation, and four heads. A hyperbolic tangent operation in the range [-3, 3] is applied on top of the transformer encoder to ensure the transformer's output is not too large. More details about the training setting can be found in Appendix \ref{app:details}.
To account for statistical variations, each algorithm is run three times in full. All experiments are executed on a single Nvidia A100-SXM4 GPU with 40GB of RAM, using Python and the PyTorch library. The source code is available at \url{https://github.com/annaVettoruzzo/L2L_with_attention}.

\begin{figure}[tbp]
\centering
\includegraphics[width=0.95\textwidth]{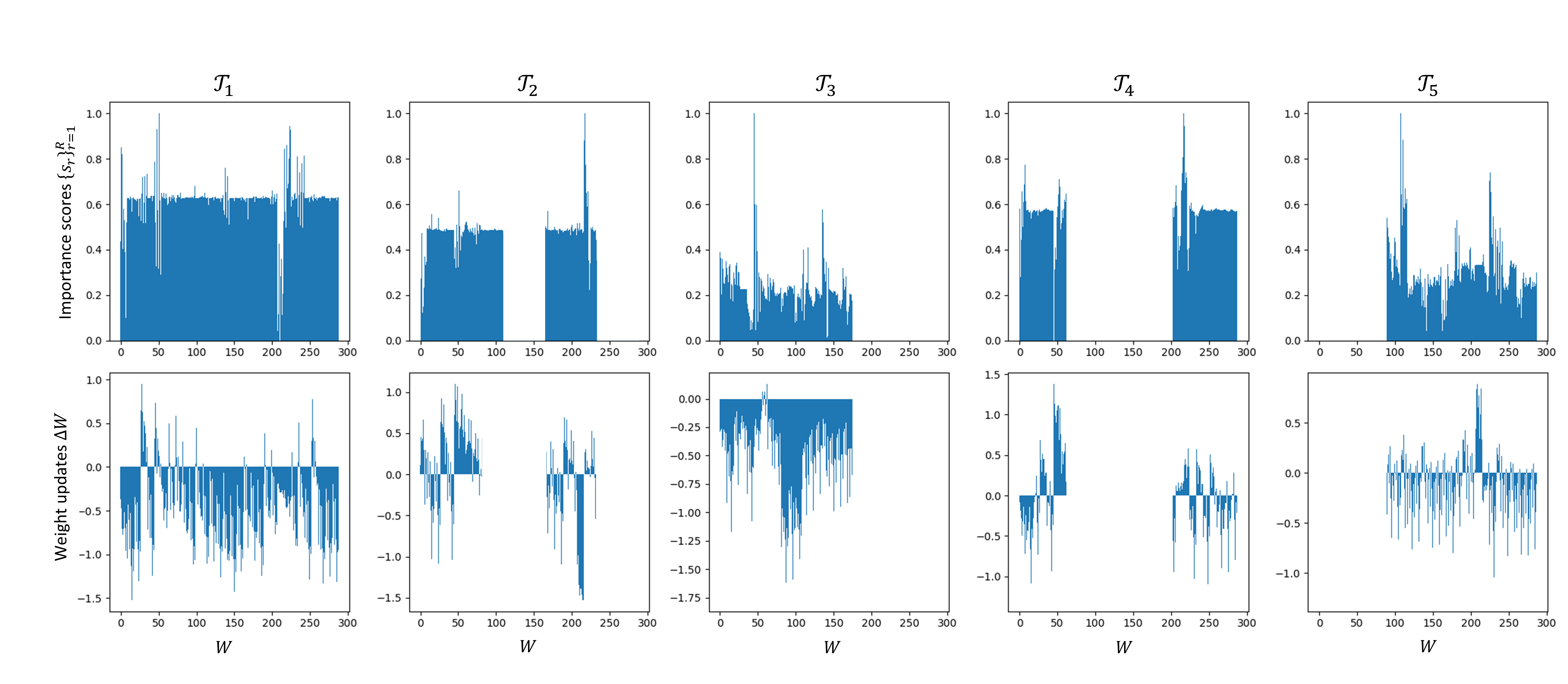}
\caption{Visualization of the importance scores $\{s_r\}_{r=1}^R$ and the weight updates learned by the meta-optimizer $g_{\psi}$ after adaptation to each task $\mathcal{T}_{i} \in \protect\vv{\mathcal{T}}$ for SplitMNIST dataset.} 
\label{fig:predictions}
\end{figure}

\subsection{Results}
First, we verify whether the meta-optimizer learns to optimize weights selectively, updating only the weights considered important for a given task (i.e., resulting in a high importance score), hence resulting in low forgetting. Figure \ref{fig:predictions} shows the importance scores given to the meta-optimizer (top), and the corresponding weight updates learned by the meta-optimizer for each task in the training stream (bottom) for SplitMNIST, showing a strong correlation. More experiments on the influence of the importance scores in the proposed approach can be found in Appendix \ref{app:ablation}.

The full performance results are presented in Table \ref{tab:mnist_results} for SplitMNIST/RotatedMNIST and in Table \ref{tab:cifar_results} for SplitCIFAR-100. To highlight the effectiveness of the proposed approach in balancing the different metrics, Figure \ref{fig:pareto} shows a Pareto plot illustrating the average BWT versus FWT across all datasets.
On average, our proposed approach significantly outperforms most other CL techniques in both forward and backward transfer. It is competitive with Sparse-MAML on BWT but significantly outperforms it in terms of FWT. Indeed, the primary goal of this paper is to introduce an approach that not only prevents forgetting previous knowledge, as indicated by the BWT metric, but also ensures robust performance on future tasks while leveraging only a small set of labeled data, which is particularly noticeable in the FWT metric, where the proposed approach outperforms all other baselines by a significant margin. In terms of ``Avg accuracy" and BWT, TIL methods (such as EWC, SI, La-MAML, Sparse-MAML, and WSN) show superior results on SplitMNIST and SplitCIFAR-100. However, these methods require additional information during training and evaluation, as discussed in Section \ref{sec:relatedworks}. Specifically, TIL methods use a task identifier $t_m$ to implement a multi-head approach, enhancing prediction outcomes, especially in simpler datasets like SplitMNIST. Nevertheless, when applied to more complex datasets, such as RotatedMNIST and SplitCIFAR-100, TIL methods show a large negative BWT, indicating their inability to retain past knowledge. An exception is WSN, which has a BWT close to 0 on SplitMNIST and SplitCIFAR-100. Indeed, as stated in the original paper \citep{kang22b}, WSN is almost immune to catastrophic forgetting as each task is associated with a distinct subnetwork model that does not interfere with others. However, this poses a challenge as the number of tasks increases and the network exhausts its capacity to allocate new weights, as observed in RotatedMNIST. The low BWT, in this case, is also influenced by the use of a single-head classifier, given that the structure of the problem, i.e., classifying digits from 0 to 9, remains the same for all tasks. 
CIL methods like iCaRL and DER++ show lower performance on SplitMNIST and SplitCIFAR-100. These methods infer the task identity and expand the classifier head as new classes are encountered. This approach is more challenging compared to TIL and the method proposed in this paper. Yet, when considering the RotatedMNIST dataset, where CIL methods behave similarly to the proposed approach since the classes are all the same across tasks, our method still shows better backward and forward transfer. Specifically, while iCaRL cannot be extended to this scenario (thus not included in Table \ref{tab:mnist_results}), the proposed approach shows a better backward and forward transfer compared to DER++, highlighting the efficacy of the meta-optimizer to extract context from the few input data provided for each task and learn an optimization strategy tailored to that task.
\begin{table} [tbp]
\centering
\caption{Performance comparison on (a) SplitMNIST and RotatedMNIST and (b) SplitCIFAR-100. The baseline methods are organized into TIL methods, CIL methods, and the proposed approach. Results report the mean and std across three different runs of all the algorithms for Average accuracy, BWT and FWT.} \label{tab:results}
\begin{subtable}{\linewidth}
\centering
\begin{adjustbox}{width=\textwidth,center}
\begin{tabular}{ lccc|ccc} 
\hline
\textbf{Method} & \multicolumn{3}{c}{\textbf{SplitMNIST}} & \multicolumn{3}{c}{\textbf{RotatedMNIST}} \\
\cline{2-4} \cline{5-7}
& Avg accuracy & BWT & FWT & Avg accuracy & BWT & FWT \\
\hline
EWC & $\mathbf{99.76 \pm 0.04}$ & $-0.03 \pm 0.01$ & $68.40 \pm 4.48$ & $53.32 \pm 1.95$ & $-38.86 \pm 0.49$ & $54.94 \pm 0.50$ \\
SI & $99.82 \pm 0.00$ & $\mathbf{-0.01 \pm 0.01}$ & $70.62 \pm 4.01$ & $45.46 \pm 1.92$ & $-48.97 \pm 0.43$ & $55.42 \pm 0.36$\\
La-MAML & $99.35 \pm 0.18$ & $-0.24 \pm 0.13$ & $68.67 \pm 4.45$ & $66.45 \pm 1.82$ & $-23.78 \pm 1.03$ & $52.95 \pm 0.52$ \\
Sparse-MAML& $93.54 \pm 3.00$ & $-5.49 \pm 2.93$ & $82.56 \pm 2.53$ & $12.41 \pm 1.61$ & $-15.81 \pm 2.30$ & $16.03 \pm 0.15$ \\
WSN &$99.74 \pm 0.05$ & $-0.09 \pm 0.03$ & $70.22 \pm 3.56$ & $45.28 \pm 2.52$ & $-48.46 \pm 1.14$ & $56.04 \pm 0.55$\\
\hline
iCaRL& $84.85 \pm 0.20$ & $-8.63 \pm 0.11$ & $36.85 \pm 0.04$ & $-$ & $-$ & $-$\\
DER++ & $84.61 \pm 2.61$ & $-11.03 \pm 2.58$ & $68.14 \pm 1.45$& $\mathbf{73.67 \pm 0.75}$ & $-18.91 \pm 0.70$ & $55.65\pm 0.90$ \\
\hline
Proposed & $92.78 \pm 0.66$ & $-6.74 \pm 1.52$ & $\mathbf{90.40 \pm 0.26}$ & $62.09 \pm 1.78$ & $\mathbf{-1.70 \pm 1.08}$ & $\mathbf{66.07 \pm 0.61}$\\
\hline
\end{tabular}
\end{adjustbox}
\caption{Results on SplitMNIST and RotatedMNIST}
\label{tab:mnist_results}
\end{subtable}

\begin{minipage}[t]{0.5\textwidth}
\vspace{5pt} 
\begin{adjustbox}{width=0.96\textwidth}
\begin{subtable}{\linewidth}
\begin{tabular}{lccc} 
\hline
\textbf{Method} & \multicolumn{3}{c}{\textbf{SplitCIFAR-100}} \\
\cline{2-4} 
& Avg accuracy & BWT & FWT\\
\hline
EWC & $52.49 \pm 1.62$ & $-29.66 \pm 1.97$ & $43.29 \pm 0.12$\\
SI & $55.84 \pm 0.83 $ & $-25.32 \pm 1.26$ & $42.98 \pm 0.50$\\
La-MAML & $52.64 \pm 1.60$ & $-22.59 \pm 0.84$ & $41.10 \pm 0.37$ \\
Sparse-MAML& $37.68 \pm 3.57$ & $-0.82 \pm 1.30$ & $34.39 \pm 1.61$\\
WSN & $\mathbf{82.88 \pm 0.13}$ & $\mathbf{-0.22\pm 0.10}$ & $44.47 \pm 0.95$\\
\hline
iCaRL& $20.59 \pm 1.58$ & $-12.97 \pm 0.92$ & $10.67 \pm 0.63$\\
DER++ & $41.32 \pm 0.41$ & $-36.11 \pm 0.61$ & $39.62 \pm 1.36$ \\
\hline
Proposed & $49.96 \pm 2.33 $ & $-14.95 \pm 3.16$ & $\mathbf{49.95 \pm 0.31}$ \\
\hline
\end{tabular}
\caption{Results on SplitCIFAR-100}
\label{tab:cifar_results}
\end{subtable}
\end{adjustbox}
\end{minipage}
\hfill
\begin{minipage}[t]{0.4\textwidth}
\vspace{0pt} 
\includegraphics[width=1\textwidth]{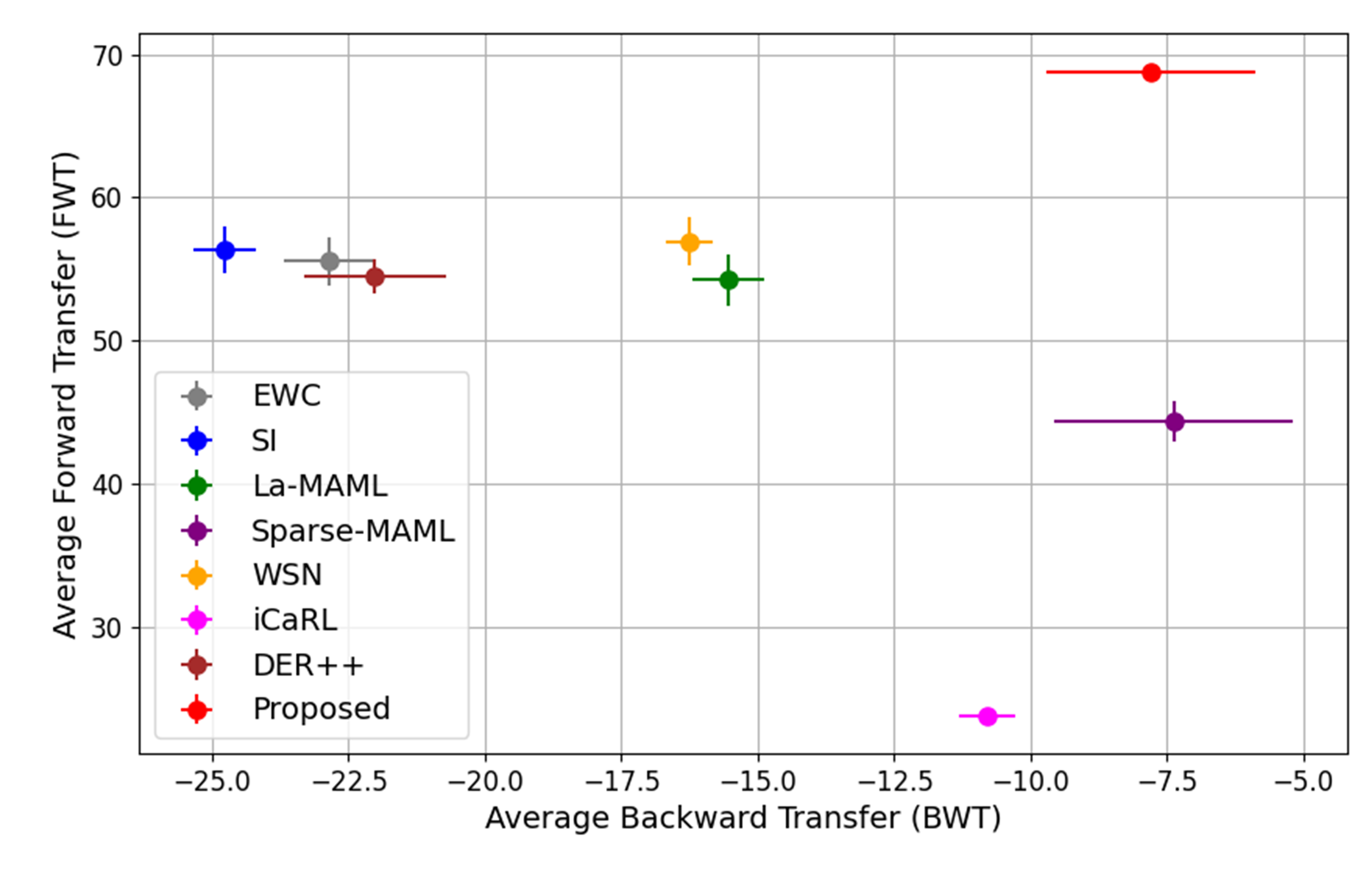}
\vspace{-0.8cm}
\captionof{figure}{Pareto plot of average BWT vs. FWT}
\label{fig:pareto}
\end{minipage}
\end{table}

\begin{figure}[t]
\centering
\subfloat[][Average accuracy] 
{\includegraphics[width=.33\textwidth]{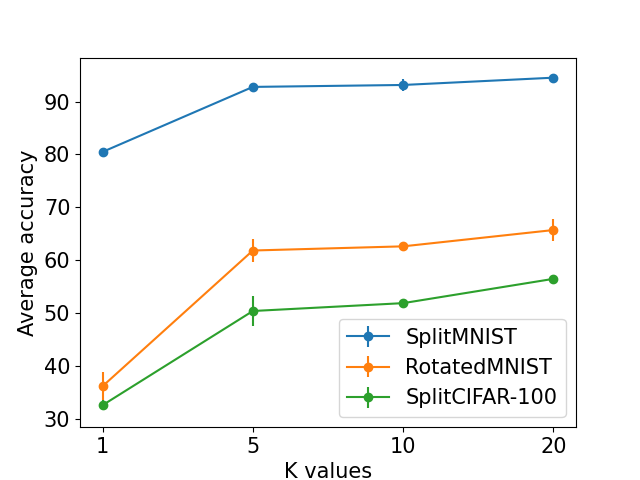}\label{fig:accuracy_kvalues}}
\subfloat[][BWT]
{\includegraphics[width=.33\textwidth]{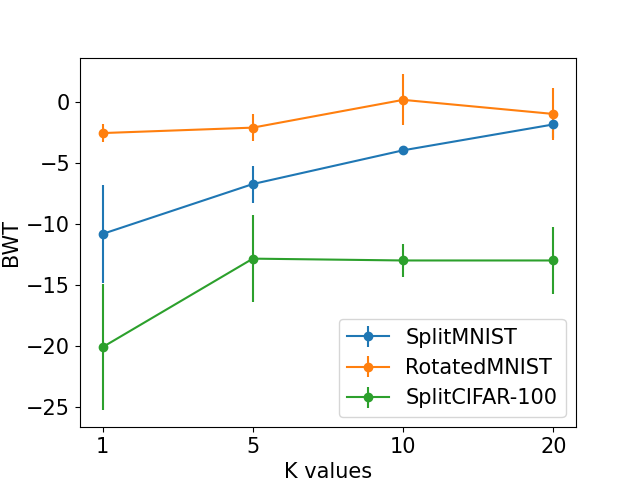}\label{fig:bwt_kvalues}}
\subfloat[][FWT]
{\includegraphics[width=.33\textwidth]{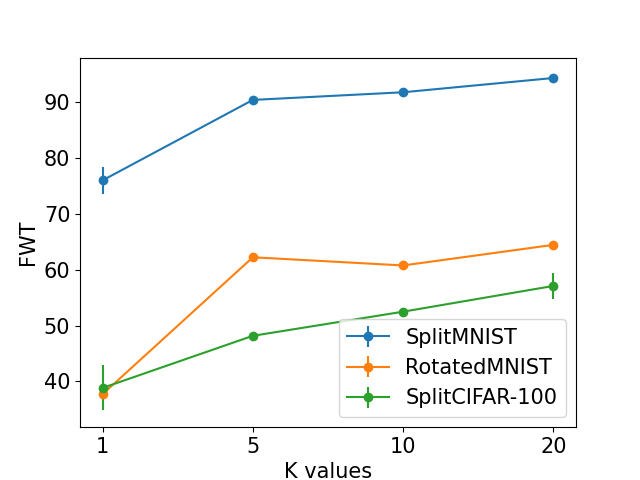}\label{fig:fwt_kvalues}} 
\caption{Visualization of the three metrics (average accuracy, BWT, FWT) varying the number of labeled examples in the support set of each task from $K=1$ to $K=20$, for all three datasets.}
\label{fig:kvalues}
\end{figure}
Even better results can be achieved by increasing the number of labeled examples in the support sets, allowing the model to better capture task-specific information and thus better adapt the meta-optimizer to predict weight updates specifically tailored for the task at hand. This is shown in Figure \ref{fig:kvalues}, where the three performance metrics are computed for the three datasets when training the proposed approach with $K=1, 5, 10,$ and $20$ examples.

\begin{figure}[!t]
\begin{minipage}[t]{0.58\textwidth}
\begin{table}[H]
\caption{Comparison between the proposed approach and MAML \citep{finn2017model} The primary distinction between these two methods is the use of the meta-optimizer in the proposed approach to learn a task-specific update rule, as opposed to directly optimizing all the weights of the classifier model using gradient descent, as in MAML. Results report the mean and std across three different runs of all the algorithms for ``Avg accuracy", BWT and FWT.}
\label{tab:ablation}
\centering
\begin{adjustbox}{width=\textwidth,center}
\begin{tabular}{ lccc} 
\hline
&& MAML & Proposed \\
\hline
\multirow{3}{*}{SplitMNIST} & \multicolumn{1}{l}{Avg accuracy} & \multicolumn{1}{c}{$86.11 \pm 1.30$}  &\multicolumn{1}{c}{$\mathbf{92.78 \pm 0.66}$} \\
                                 & \multicolumn{1}{l}{BWT} & \multicolumn{1}{c}{$-14.28 \pm 1.36$}  &\multicolumn{1}{c}{$\mathbf{-6.74 \pm 1.52}$} \\
                                 & \multicolumn{1}{l}{FWT} & \multicolumn{1}{c}{$88.72 \pm 0.54$}  &\multicolumn{1}{c}{$\mathbf{90.40 \pm 0.26}$} \\\hline
\multirow{3}{*}{RotatedMNIST} & \multicolumn{1}{c}{Avg accuracy} & \multicolumn{1}{c}{$\mathbf{65.65 \pm 1.36}$}  & \multicolumn{1}{c}{$62.09 \pm 1.78$}\\
                                 & \multicolumn{1}{l}{BWT} & \multicolumn{1}{c}{$-19.46 \pm 0.98$}  & \multicolumn{1}{c}{$\mathbf{-1.70 \pm 1.08}$}\\
                                 & \multicolumn{1}{l}{FWT} & \multicolumn{1}{c}{$\mathbf{67.00 \pm 0.37}$}  & \multicolumn{1}{c}{$66.07 \pm 0.61$} \\\hline
\multirow{3}{*}{SplitCIFAR-100} & \multicolumn{1}{l}{Avg accuracy} & \multicolumn{1}{c}{$46.15 \pm 0.96$} & \multicolumn{1}{c}{$\mathbf{49.96 \pm 2.33 }$ } \\
                                 & \multicolumn{1}{l}{BWT} & \multicolumn{1}{c}{$-36.22 \pm 2.82$} & \multicolumn{1}{c}{$\mathbf{-14.95 \pm 3.16}$ } \\
                                 & \multicolumn{1}{l}{FWT} & \multicolumn{1}{c}{$49.72 \pm 0.45$} & \multicolumn{1}{c}{$\mathbf{49.95 \pm 0.31}$} \\\hline
\end{tabular}
\end{adjustbox}
\end{table}
\end{minipage}
\hfill
\begin{minipage}[t]{0.4\textwidth}
\begin{figure}[H]
\centering
\includegraphics[width=1\textwidth]{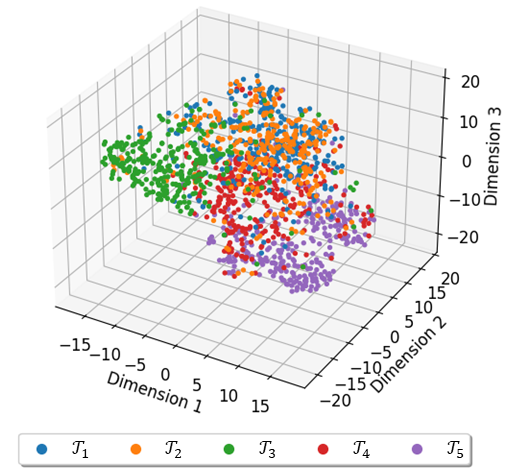}
\caption{Visualizaton of the optimized weights at test time for all  SplitMNIST tasks $\{\mathcal{T}_m\}_{m=1}^M$.}
\label{fig:weights3d}
\end{figure}
\end{minipage}
\end{figure}

To assess the meta-optimizer's capability in effectively learning an optimization strategy tailored for each task, we perform an ablation study comparing MAML with the proposed approach, as shown in Table \ref{tab:ablation}. The results indicate that the proposed approach outperforms MAML in both SplitMNIST and SplitCIFAR-100. Only on the RotatedMNIST dataset, both MAML and the proposed approach yield comparable results. To better understand this, we can examine the test accuracy for different tasks during training. While MAML achieves better test accuracy on the last-seen task, it suffers from catastrophic forgetting (i.e., high BWT). In contrast, the test accuracy obtained with the proposed approach is lower for the last-seen task, but the BWT is significantly improved. This lower accuracy can be attributed to the limited network capacity, which constrains the classifier's ability to diversify weight updates across different tasks. However, while we could not experiment with larger network architectures due to memory constraints, that would likely address this limitation.

Finally, Figure \ref{fig:weights3d} illustrates the ability of the model to learn diverse weight updates on small training samples using the same model architecture. This plot visualizes the optimized weights $W$ for the different tasks in SplitMNIST by applying tSNE \citep{van2008visualizing} to reduce the data dimensionality.
As expected, the weights are optimized differently for each task, showcasing the ability of the meta-optimizer to effectively recognize task differences without requiring a task identifier. Moreover, certain weight values remain constant across different tasks, allowing the classifier model to reuse previously acquired knowledge in solving similar tasks.

\section{Conclusion} \label{sec:conclusions}
This paper introduces a novel approach to address the challenges of CL by integrating meta-learning techniques. The primary contribution is the introduction of a meta-learned transformer-based optimizer, designed to learn an update strategy that enhances the capability of machine learning models to continuously learn while retaining knowledge over time. 
Notably, the proposed approach does not require a memory buffer to store information from past tasks. It can scale to long streams of tasks since it doesn't add new parameters to the model. It could also scale to larger images by using larger classifiers and focusing on updating the last layer. Indeed, the complexity of the meta-model primarily depends on the number of weights to optimize rather than the input data. Furthermore, using a meta-learned optimizer eliminates the need to manually select hyperparameters for the classifier network, which is critical in dynamic CL environments. Through comprehensive evaluations on benchmark datasets, the results demonstrate the effectiveness of the proposed approach. Leveraging only a limited set of labeled data, the method achieves strong performance in terms of both backward and forward transfer. The meta-optimizer effectively predicts task-specific weight updates, showcasing its potential in real-world applications where adapting to new tasks with limited labeled data is crucial.

Finally, the pioneering use of a transformer model as a meta-optimizer in the context of CL opens up avenues for future research to further explore its capabilities. In particular, it would be interesting to meta-learn larger transformer models on longer streams, which should improve both BWT and FWT as it can meta-learn over more tasks, or to combine it with episodic memory buffers to improve BWT. Further works could extend the $N$-way classification approach detailed in this paper to accommodate a dynamic classifier head capable of predicting all encountered classes, as in CIL. Finally, given that the optimizer works with batches of data as input, the proposed approach could also be extended to online CL scenarios.

\section*{Acknowledgements}
This work was supported by the EU’s Horizon Europe research and innovation program under grant agreement No. 952215 (TAILOR).

\bibliography{collas2024_conference}
\bibliographystyle{collas2024_conference}

\clearpage
\appendix

\section{Appendix}

\subsection{Experimental details} \label{app:details}

For training the models, we use the same dataset splits described in \cite{jin_gradient-based_2021}. During training, we select $K=5$ samples per class for the support set and we utilize a maximum of $100$ examples per class for the query set. For each task, the model is trained for $10^5$ steps. We set the number of adaptation steps, $Q$ in Algorithm \ref{algo2}, to three for both SplitMNIST and RotatedMNIST, and to five for SplitCIFAR-100. However,  during test time, we perform $50$ adaptation steps to achieve the reported results. The SGD optimizer is used in the \emph{Adapt\&Optimize} algorithm (Algorithm \ref{algo2}) with a learning rate of $10^{-3}$, while the Adam optimizer is used for updating the parameters $\theta$, in line \ref{line:update} of Algorithm \ref{algo1}, with a learning rate of $10^{-4}$ and a weight decay value of $10^{-5}$. 
To update the transformer's parameter, $\psi \in \theta$, we utilize a learning rate schedule comprising a warmup phase of $3000$ steps followed by linear decay, as recommended by \cite{vaswani2017attention}. We also employ an L2 regularizer to avoid exploding gradients.

Due to the large variety of baseline methods selected for performance comparison, we try to be as consistent as possible with the methodology proposed in the original papers, utilizing the default values for the parameters specific to each method. For the test setting, we adopt the same approach used in the original papers \citep{kirkpatrick2017overcoming, zenke2017continual, gupta2020look, von2021learning, kang22b, rebuffi2017icarl, buzzega2020dark}, meaning that only in La-MAML \citep{gupta2020look} and Sparse-MAML \citep{von2021learning} the model is fine-tuned on a subset of the test task, before making predictions on the remaining test data. Results do not show a considerable improvement for these methods either considering the BWT and the FWT.
While fine-tuning for enough time on the new task could improve the model performance, mostly considering the FWT metric, we hypothesize that fine-tuning only for a few steps (e.g., $50$ in our experiments) does not influence the final results.

Two distinct classifier architectures are used in the experiments, with similar architectures as the ones proposed in La-MAML \citep{gupta2020look}, Sparse-MAML \citep{van2008visualizing}, and WSN \citep{kang22b}, which are the closest to our approach. For the SplitMNIST and RotatedMNIST datasets, we utilize a simple classifier, including a convolutional layer with 32 filters of size 3, followed by batch normalization, LeakyReLU, max pooling with size 2, and a classifier head. A 3-layer CNN is used for SplitCIFAR-100 comprising three convolutional layers with 100 filters and a kernel size of 3, also followed by batch normalization, LeakyReLU, and max pooling, and concludes with a classifier head. 
The architecture of the meta-optimizer comprises a task encoder (part A), a feature extractor (part B), and a transformer encoder (part C), as described in Section \ref{subsec:metaoptim}. The task encoder includes a pre-trained feature extractor, followed by an averaging layer to compute the mean of the extracted features, yielding a single vector representing the support set of the task. For SplitMNIST and RotatedMNIST experiments, the task encoder is pre-trained on the SVHN dataset, and it comprises two hidden layers with 100 neurons. Conversely, for SplitCIFAR-100 experiments, the convolutional part of VGG-11, pre-trained on ImageNet, is employed. The second component of the meta-optimizer (i.e., part B) is a shallow network consisting of a single hidden layer with 16 neurons. Finally, the transformer encoder consists of two stacked transformer encoder layers, each incorporating a multi-head self-attention block with four attention heads and a feed-forward network with a size of $64$. We add positional embeddings to the input features using a learnable 1D encoding, similar to \cite{dosovitskiy2020image}. The transformer layers are trained using the GELU activation function and dropout regularization with a probability of 0.2. Additionally, we apply a hyperbolic tangent operation within the range $[-3, 3]$ on top of the transformer encoder to ensure the output does not become too large.

\subsection{Ablation studies on the meta-optimizer} \label{app:ablation}

Due to the complexity of the meta-optimizer, we conduct several ablation studies to verify the significance of each component in achieving the final performance. These ablation studies are performed by removing one component at a time while keeping the other parts consistent, as outlined below:
\begin{itemize}
    \item The pre-trained task encoder $g_A$ in Figure \ref{fig:optimizer} is replaced by a simpler model comprising of a flattening operation, an averaging layer over the $KN$ examples in the input support set, and a simple linear layer that transforms the average representation into a size suitable for the transformer model.
    \item The feature extractor $g_B$ in Figure \ref{fig:optimizer} is modified with a simple linear layer that transforms the input weights into a size suitable for the transformer encoder.
    \item The transformer encoder is replaced by an LSTM network with two LSTM layers with the same hidden size as the transformer encoder, followed by a linear layer and a hyperbolic tangent operation in the range $[-3, 3]$, as used with the transformer encoder.
    \item The importance scores are ablated by simply removing them from the proposed approach.    
\end{itemize}
The results in Table \ref{tab:ablation_optimizer} indicate that while the task encoder and the feature extractor contribute marginally to the final performance, the transformer encoder and the importance scores are necessary for the proposed approach. This finding aligns with the expectations, as the transformer encoder, unaffected by long dependency issues owing to the non-sequential input processing, plays a crucial role in learning to optimize for CL \citep{chen2022learning}. Additionally, the self-attention mechanism and the positional embeddings provide even more information about the relationship between the input weights, which is fundamental in the proposed approach. The other essential component is the importance scores. Containing information related to the gradient of the loss function they help the transformer encoder to learn to optimize only the classifier weights that are more important for the current task. This mitigates catastrophic forgetting of the model. Notably, removing the importance scores results in a significant decrease in the BWT metric compared to the proposed approach, underscoring their pivotal contribution. It is also interesting to notice that removing the importance scores causes a big drop in the FWT metric, suggesting that learning to optimize all the model weights ends up in a model too tailored to the current task that cannot easily be generalized to new tasks.

\begin{table*}[tbp]
\caption{Ablation studies on the different components of the meta-optimizer for SplitMNIST and SplitCIFAR-100. The first column shows the ablated component and results show the average accuracy (``Avg Acc"), BWT and FWT metric as described in Section \ref{sec:ds_and_methods} . }
\label{tab:ablation_optimizer}
\centering
\begin{tabular}{ l|ccc|ccc} 
\hline
\textbf{Ablated component}& \multicolumn{3}{c}{\textbf{SplitMNIST}} & \multicolumn{3}{c}{\textbf{SplitCIFAR-100}} \\
 & Avg Acc & BWT & FWT & Avg Acc & BWT & FWT \\
\hline
Task encoder & $93.18$& $-8.55$ & $88.24$& $47.71$& $-16.31$& $48.37$\\
Feature extractor & $92.56$& $-10.56$ & $82.18$& $47.01$& $-17.24$& $43.76$\\
Transformer encoder &$70.80$ & $-27.22$ & $90.52$ & $25.45$& $-2.22$& $25.59$\\
Importance scores & $84.26$ & $-16.15$ & $80.40$ & $43.17$& $-18.90$& $41.78$\\
\hline
\end{tabular}
\end{table*}

\subsection{Transfer learning baseline}

\begin{figure}[!t]
\begin{minipage}[t]{0.65\textwidth}
\begin{figure}[H]
\centering
\includegraphics[width=0.8\textwidth]{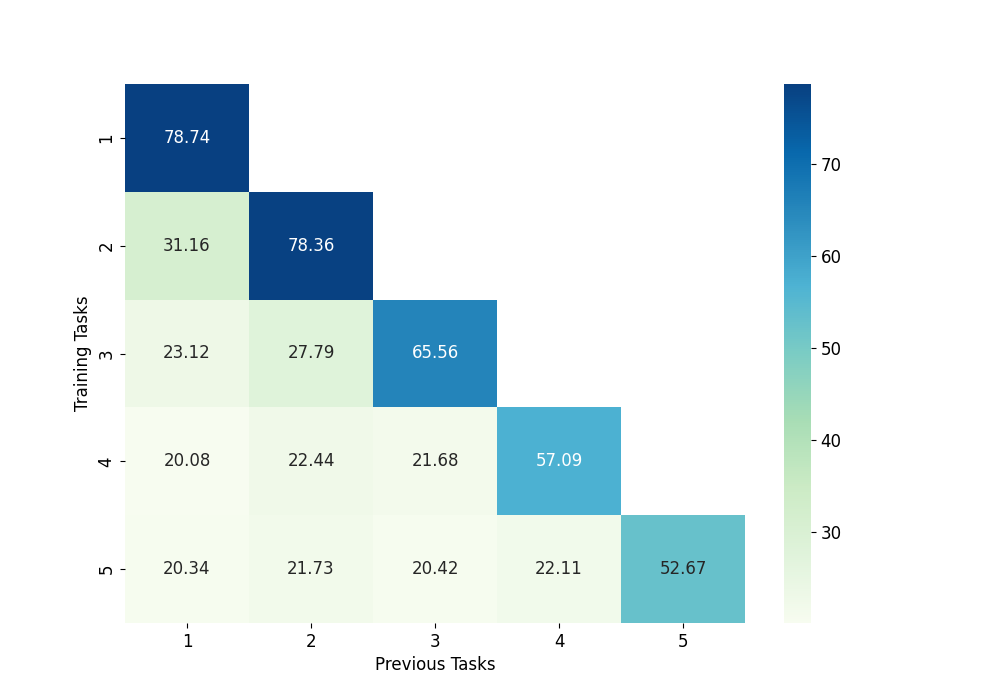}
\caption{Accuracy matrix at test time for the transfer learning baseline with SplitCIFAR-100. Each cell of the matrix $a_{ij}$ represents the accuracy of task $\mathcal{T}_j$ when the last task seen by the model is $\mathcal{T}_i$.}
\label{fig:trlearning}
\end{figure}
\end{minipage}
\hfill
\begin{minipage}[t]{0.35\textwidth}
\begin{table}[H]
\vspace{1.8cm}
\caption{Results for the transfer learning baseline with SplitCIFAR-100 dataset using a VGG11 model pre-trained on ImageNet and a classifier model on top of it.}
\label{tab:trlearning}
\centering
\begin{tabular}{ lc} 
\hline
\multicolumn{2}{c}{\textbf{Transfer learning} }\\
\hline
Avg accuracy &$27.45$\\
BWT & $-50.74$\\
FWT &$41.97$\\
\hline
\end{tabular}
\end{table}
\end{minipage}
\end{figure}

A simple transfer learning baseline is implemented to verify that solely pre-training the task encoder $g_A$ in the meta-optimizer (see Figure \ref{fig:optimizer}) cannot directly address the problem outlined in this paper.
Specifically, we concatenate the task encoder pre-trained on ImageNet with the classifier model used with SplitCIFAR-100, and we fine-tune only the classifier on the support set of the tasks. To ensure fairness with other baseline methods proposed in this paper, we fine-tune the classifier model for $10^5$ steps using the Adam optimizer with a learning rate set to $10^{-4}$ and a weight decay value of $10^{-5}$ for the current task observed by the model. Subsequently, we evaluate it on previous and future tasks by fine-tuning the model with only $50$ adaptation steps.
Results in Figure \ref{fig:trlearning} and Table \ref{tab:trlearning} demonstrate that while the model achieves high performance on the last seen task (in the diagonal of the matrix in Figure \ref{fig:trlearning}), it fails to generalize to previous or future tasks, as highlighted by the low BWT and FWT metrics. This outcome is expected since mere application of transfer learning is inadequate to effectively retain previous knowledge and generalize to future tasks.


\subsection{Limitations and solutions}
While our approach of meta-learning a meta-optimizer using attention demonstrates promising results in addressing CL challenges, it is important to acknowledge some inherent limitations. Firstly, the tight coupling between the meta-optimizer and the classifier imposes constraints on the applicability of the trained meta-optimizer to novel scenarios requiring different classifier architectures. The meta-optimizer is explicitly designed to work well for the classifier it was trained for, necessitating retraining or fine-tuning for different classifier architectures. We have demonstrated in Appendix \ref{app:ablation} that the feature extractor is not a primary contributor to the model performance. Therefore simply replacing its architecture with a more appropriate one for the number of weights $W = \{w_r\}_{r=1}^R$ that need to be learned to optimize would be sufficient to extend the proposed approach to more complex classifier architectures. Another potential issue that might arise with a deeper classifier is the scalability of the proposed approach, as the size of the input weights to the meta-optimizer increases substantially, leading to computational overheads. One potential mitigation strategy involves focusing on optimizing only the last layers, as they encode more task-specific information compared to earlier layers \citep{ramasesh2020anatomy}. 
Furthermore, the computational demands associated with second-order derivatives in meta-learning, coupled with training times and data requirements for transformers, pose practical challenges for training such a model. Lastly, the fixed size of the output classifier head limits adaptability in scenarios with varying numbers of classes, such as in CIL.  While this is not the scope of the proposed approach, future work aims to extend it to accommodate a dynamic classifier head capable of predicting all encountered classes.

\subsection{Default notation}
\bgroup
\def\arraystretch{1.5}
\begin{tabular}{p{2in}p{3.25in}}
$\mathcal{T}_m $ & A task \\
$\vv{\mathcal{T}}=\{\mathcal{T}_m\}_{m=1}^M$ & A stream of tasks \\
$p(\mathcal{T})$ & A task distribution \\
$p_m(x,y) $ & A data distribution \\
$D_{m}^{train}$ & Training data sampled from $\mathcal{T}_m$ \\
$D_{m}^{test}$ & Test data sampled from $\mathcal{T}_m$ \\
$D_{i,m}$ & A batch of data\\
$D_{i,m}^{(sp)}$ & A support set of data\\
$D_{i,m}^{(qr)}$ & A query set of data\\
$\{x_k\}_{k=1}^{KN} \in D_{i,m}^{(sp)}$ & Data in the support set \\

$K$ & K-shot classification \\
$N$ & N-way classification \\

$f_{\mu, W}$ & Classifier network \\
$g_\psi$ & Meta-optimizer network\\
$\psi$ & Parameters of the meta-optimizer\\
$\mu$ & Meta-learned parameters of the classifier network \\
$W = \{w_r\}_{r=1}^R$ & Set of classifier's parameters to meta-optimize\\
$R$ & Number of classifier's parameters to meta-optimize\\
$\theta = \{\mu, W\}$ & Parameters of the classifier network \\

$\{s_r\}_{r=1}^R$ & Importance scores\\
$\Delta W = \{\Delta w_r\}_{r=1}^R$ & Weight updates for the meta-optimized weights\\
$\theta'$ & Updated parameters with a fixed optimizer \\
$\tilde{W}$ & Updated parameters with the meta-optimizer \\
$\alpha, \beta$ & Learning rates \\
$Q$ & Adaptation steps \\
\end{tabular}
\egroup

\end{document}